\pdfoutput=1
\documentclass[10pt,twocolumn,letterpaper]{article}

\usepackage{wacv}
\usepackage{times}
\usepackage{epsfig}
\usepackage{graphicx}
\usepackage{amsmath}
\usepackage{amssymb}
\usepackage{subcaption}
\usepackage[accsupp]{axessibility}

\wacvfinalcopy 

\ifwacvfinal
\usepackage[breaklinks=true,bookmarks=false]{hyperref}
\else
\usepackage[pagebackref=true,breaklinks=true,colorlinks,bookmarks=false]{hyperref}
\fi

\begin{document}

\title{Semantic Network Interpretation}

\author{Pei Guo \quad Ryan Farrell\\
Brigham Young University\\
{\tt\small \{peiguo, farrell\}@cs.byu.edu}
}

\maketitle
\thispagestyle{empty}

\begin{abstract}
Network interpretation as an effort to reveal the features learned by a network remains largely visualization-based.
In this paper, our goal is to tackle semantic network interpretation at both filter and decision level. For filter-level interpretation, we represent the concepts a filter encodes with a probability distribution of visual attributes. The decision-level interpretation is achieved by textual summarization that generates an explanatory sentence containing clues behind a network's decision. A Bayesian inference algorithm is proposed to automatically associate filters and network decisions with visual attributes. Human study confirms that the semantic interpretation is a beneficial alternative or complement to visualization methods. We demonstrate the crucial role that semantic network interpretation can play in understanding a network's failure patterns. More importantly, semantic network interpretation enables a better understanding of the correlation between a model's performance and its distribution metrics like filter selectivity and concept sparseness. 
\end{abstract}

\section{Introduction}
Network interpretation seeks to illuminate or expose the features that have been learned,
and its difficulty lies in the end-to-end learning of the feature extraction and classification sub-networks, which typically contain millions of parameters each. ``Debugging'' an over-confident network, one which assigns the wrong class label to an image with high confidence, can be extremely difficult, especially when adversarial noise~\cite{Goodfellow_2014_ArXiv} is added to deliberately mislead the network to the wrong conclusion. In that case a meaningful explanation is highly desirable, which contains features responsible for triggering the error, similar to the syntax error highlighting of an intelligent compiler.
A thorough understanding of the neural networks is an indispensable part for their continuous success. Network interpretation is also crucial for tasks involving humans due to legal reasons. It is therefore important to distill the knowledge learned by deep models and present it in an easy-to-understand way.

\begin{figure}[t]
  \centering
  \includegraphics[width=\linewidth]{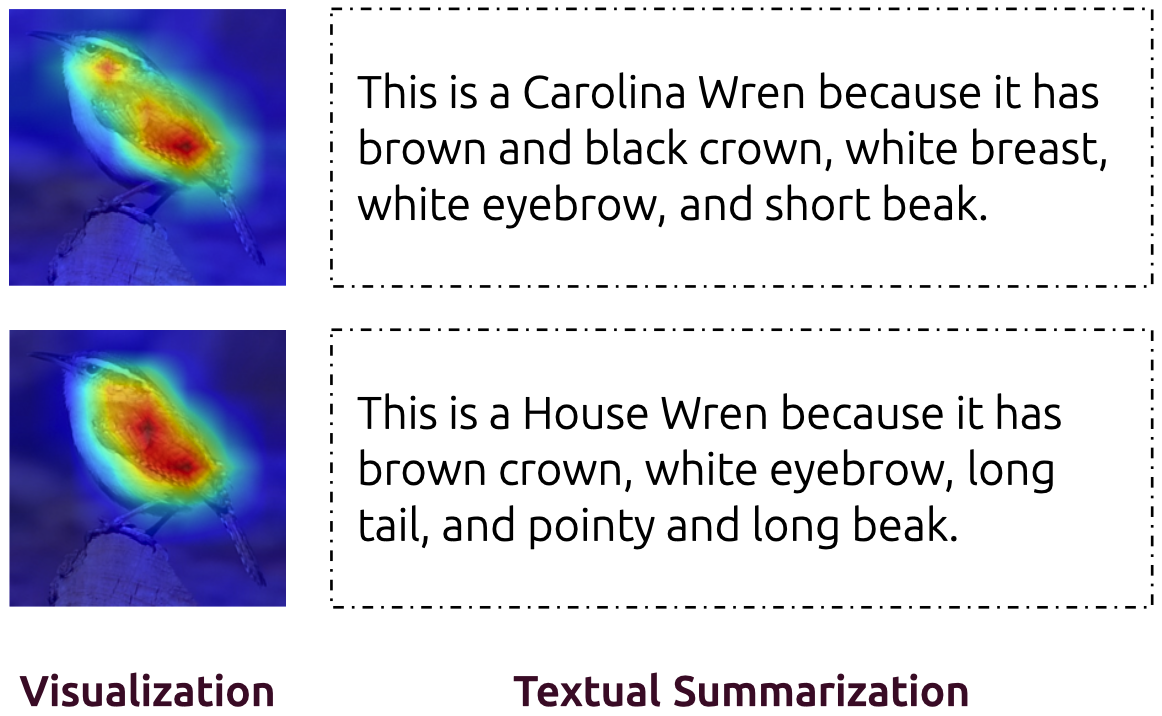}
  \caption[Semantic network interpretation]{
  Visualization methods highlight the important image region for different network decisions, but they lack semantic information and finer details compared to semantic interpretation via textual summarization.
  }
  \label{fig:overview}
\end{figure}



\begin{figure}[t]

  \centering
  \includegraphics[width=0.9\linewidth]{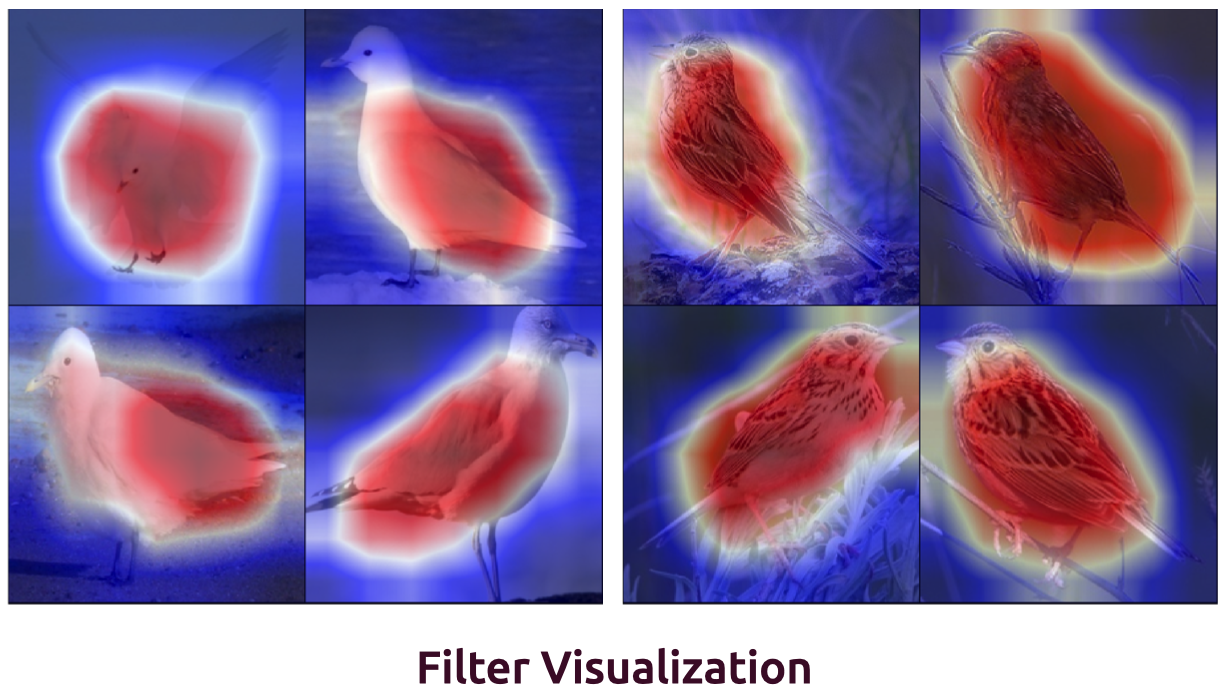}
  \caption[Examples of filter visualization]{
  Examples of filter visualization using images with maximal activation, each masked by their corresponding feature maps. One limitation of the visualization-based filter interpretation is the lacking of \textbf{diversity}: it is unable to capture the whole space of represented concepts with a limited number of data samples.}
  \label{fig:filter-vis}
\end{figure}

Most popular approaches for network interpretation are visualization-based. Filter-level interpretation (understanding the concepts a filter encodes) is often achieved by displaying the maximally activated dataset example~\cite{Zeiler_2013_ECCV} (Figure~\ref{fig:filter-vis}) or the optimized input image with image prior regulation~\cite{Nguyen_2015_CVPR,NGUYEN_2016_NIPS}. Decision-level interpretation (understanding why the network makes a decision, also called attribution) ~\cite{Zhou_2016_CVPR,Selvaraju_2017_ICCV,sundararajan2017axiomatic,springenberg2014striving,smilkov2017smoothgrad} is often achieved by highlighting a region in the image that's important for the decision-making. Despite their success at providing visual clues, pure visualization is unable to provide semantic explanation and sometimes misses detailed information, as shown in Figure~\ref{fig:overview}. Similarly, Adebayo,~\etal~\cite{adebayo2018sanity} argues  that ``reliance, solely, on visual assessment can be misleading.''

Humans, on the other hand, can justify their conclusions using natural language. For instance, a knowledgeable person looking at a photograph of a bird might say, ``I think this is an Anna's Hummingbird because it has a straight bill, and a red throat and crown. It's not a Broad-tailed Hummingbird because the latter lacks the red crown.'' This kind of textual description carries rich semantic information and is easily understandable. Semantic information is a logical medium in which to ground the interpretation of deep convolutional models, serving as a beneficial supplement for the visualization methods. 

\begin{figure}[t]
  \centering
  \includegraphics[width=0.9\linewidth]{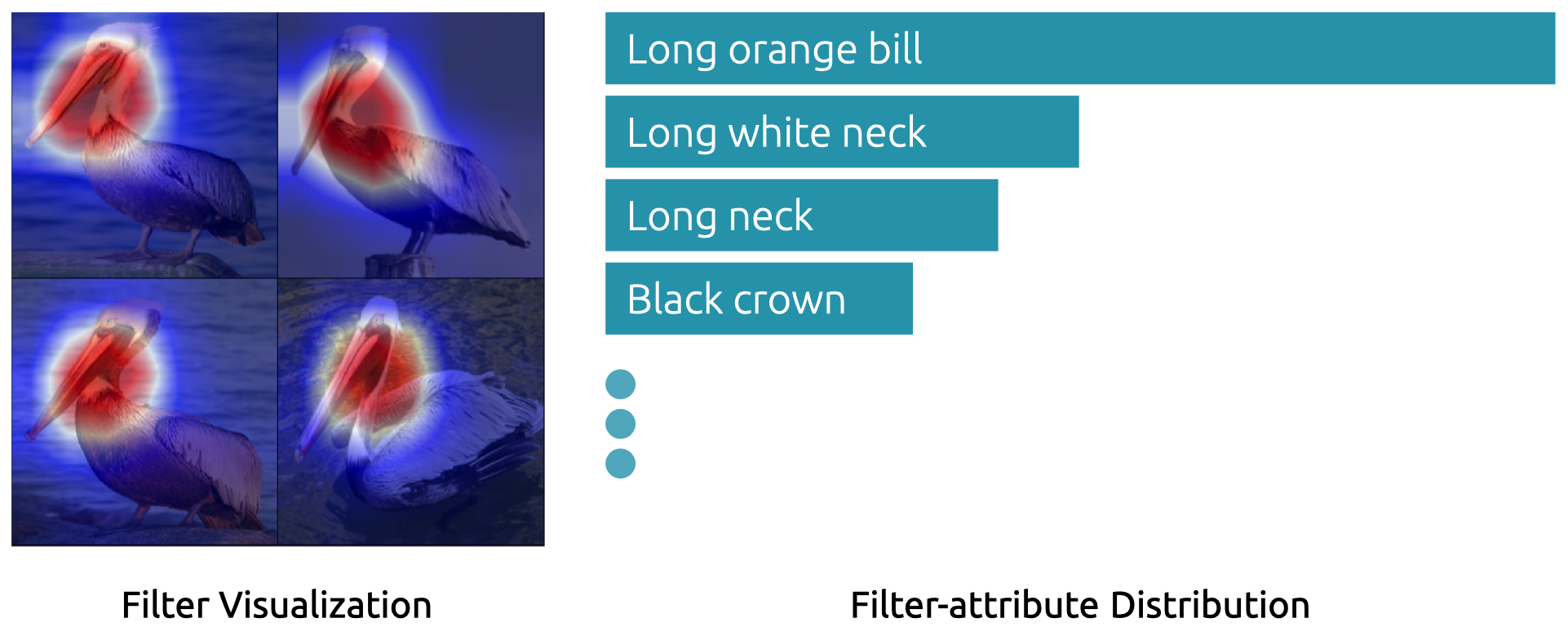}
  \caption[Filter-level semantic interpretation]{An example of filter-level semantic interpretation using filter-attribute distribution, which is a probability distribution of visual attributes that best describe the concepts encoded by a filter.}
  \label{fig:filter-att}
\end{figure}

This paper focuses on semantic network interpretation at both filter and decision level. An intuitive way for semantic filter interpretation is to assign a single concept to each filter, as did in~\cite{Bau_2017_CVPR}. However, the filter-concept relation is usually not one-to-one: a filter can represent several concepts and a concept can be encoded by multiple filters. This distributed characteristic improves a model's representation efficiency by design~\cite{hinton1984distributed}. We instead propose to represent a filter with  a conditional multinomial probability distribution, called the filter-attribute distribution (see Figure~\ref{fig:filter-att} for an example). Intuitively, an attribute $t$ is more likely to represent a filter $f$ if images containing $t$ frequently activate  filter $f$. We further tackle semantic interpretation for network decision using textual summarization. Textual summarization aims to find a list of visual attributes that the network is basing its decision on. 
A natural sentence is generated with the top attributes as supporting evidence. A direct application of  textual summarization is network debugging, generating descriptive error messages when the network's prediction is wrong, and it helps us to identify three major failure patterns for the fine-grained dataset CUB-200-2011~\cite{WahBWPB_Tech2011} (Section~\ref{sec:debugging}). 

We devise a Bayesian inference algorithm to compute the posterior probability that a filter $f$ is activated by a visual attribute $t$ as $p(t|f)$. The difference between our algorithm and a visual attribute prediction algorithm is that the later usually associates visual attributes to an image in a supervised way, but ours associates visual attributes to filters and decisions in an unsupervised way. The goal of network interpretation is not to predict the target label but to loyally reflect the internal working mechanism of a neural network. The key differences between this work and network dissection~\cite{Bau_2017_CVPR} are that we use a Bayesian algorithm to represent a filter with an attribute distribution instead of a single concept and we only leverage image-level caption annotations.

The filter-attribute distribution provides a tool to quantitatively understand how concepts are encoded by filters. Specifically, we explored the correlation between a model's performance with the distributed level of its representation. Two metrics of distributed representation are examined, namely filter selectivity and concept sparseness~\cite{Bowers_Grandmother_2011}. Filter selectivity is measured by the number of distinctive concepts a filter represents, and concept sparseness refers to the way a single concept is distributed among filters.
Understanding the correlation between a network's performance with its distributed characteristics could potentially lead to new optimization functions to train better networks. Section~\ref{sec:understanding} provides a thorough evaluation and discussion. Further more, an ablation study shows that deleting less selective filters is likely to cause more damage to a network, a contrast to our intuition. Human study shows that 41.5\% of users think textual attributes are a better medium for network interpretation than visualization, and users find 80.1\% of the top 5 attributes in the filter-attribute distributions are accurate.


\section{Related Work}

\textbf{Network interpretation  --} Two main approaches to network interpretation exist in the literature: filter-level interpretation~\cite{Erhan_2009_ICMLW,Szegedy_2013_arXiv,Mahendran_2015_CVPR,Nguyen_2015_CVPR,Google_2018_HTML,NGUYEN_2016_NIPS,Nguyen_2017_CVPR,Bau_2017_CVPR,Zhou_2014_ICLR,Yosinski_2015_ICMLW,Springenberg_2014_ICLRW,Zeiler_2013_ECCV} and decision-level interpretation (or attribution)~\cite{Simonyan_2013_ICLRW,Zhou_2016_CVPR,Selvaraju_2017_ICCV,zhou2018interpretable,HOLZINGER202128}. The goal of filter-level interpretation is to understand the features that a specific filter (also known as neurons) learns. While it is easy to directly visualize the first convolutional layer filter weights and understand the patterns they detect, it makes little sense to directly visualize deeper layer filter weights because they act as complex composite functions of lower layers' output. Early examples of filter-level interpretation include finding the maximally activated input patches~\cite{Zeiler_2013_ECCV} and visualizing the guided back propagation gradients~\cite{Springenberg_2014_ICLRW}. Some works~\cite{Nguyen_2015_CVPR} try to  synthesize visually pleasant preferred input image of each filter through back-propagation into the image space. \cite{NGUYEN_2016_NIPS} applies a generator network to generate images conditioned on maximally activating certain last-layer neurons. 
The Plug and Play paper~\cite{Nguyen_2017_CVPR} further extends~\cite{NGUYEN_2016_NIPS} to introduce a generalized adversarial learning framework for filter-guided image generation. 
Network dissection~\cite{Bau_2017_CVPR} connects each filter with predefined concepts like object, part, color, etc.

\begin{figure*}[ht]
  
  \centering
  \includegraphics[width=0.9\linewidth]{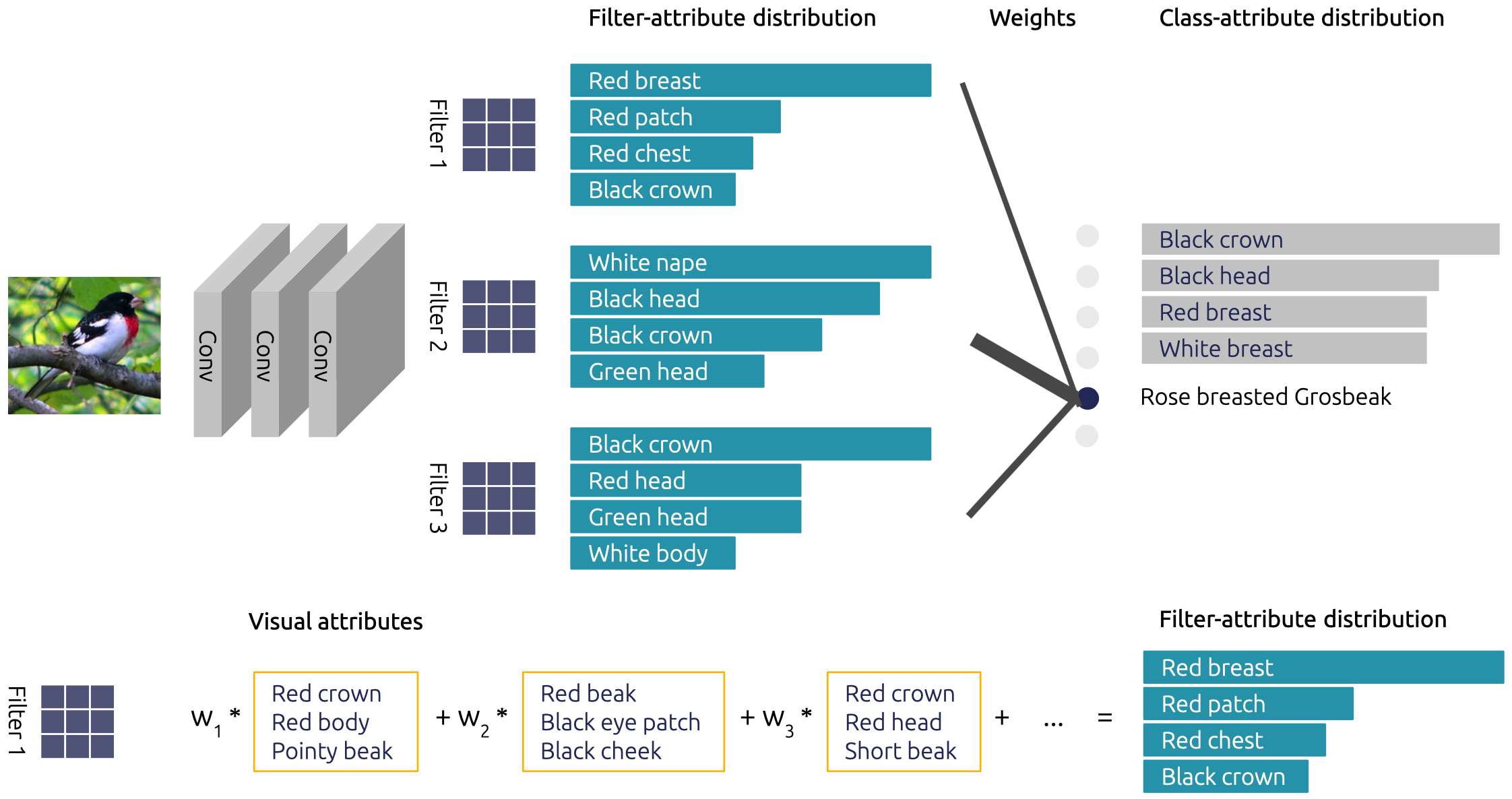}
  \caption[Algorithm overview]{An overview of the algorithms for decision-level and filter-level semantic interpretation using class-attribute distribution (top row) and filter-attribute distribution (bottom). Visual attributes in every image are weighted by the activation strength and its importance factor (TF/IDF) to generate filter-attribute distribution (section~\ref{sec:filter_apdf}). The filter-attribute distribution are re-weighted by linear layer weight and the activation strength to generate class-attribute distribution (section~\ref{sec:summarize}). }
  \label{fig:big_picture}
\end{figure*}

Attempts of decision-level interpretation  mainly focus on visualizing important image subregions by re-weighting final convolutional layer feature maps. Examples include ~\cite{Zhou_2016_CVPR, Selvaraju_2017_ICCV,sundararajan2017axiomatic,springenberg2014striving,smilkov2017smoothgrad}. However, the visualization based method only provides coarse-level information, and it remains hard to intuitively know what feature or pattern the network has learned to detect. More importantly, the holistic heat map representation is sometimes insufficient to justify why the network favors certain classes over others when the attentional maps for different classes overlap heavily. See Figure~\ref{fig:overview} for example. \cite{Zhang_2018_AAAI} proposes represent the image content and structure by knowledge graph.

\textbf{Visual attribute prediction and image captioning --} Visual attribute prediction and image captioning~\cite{Farhadi_2010_ECCV} are related but fundamentally different tasks to semantic network interpretation. Visual attribute prediction and image captioning are often supervised with the goal to approximate ground truth labels. Semantic interpretation, on the other hand, aims to loyally reflect the knowledge learned by a model in an unsupervised way. There are no ground truth labels to approximate in semantic interpretation.

We note that~\cite{hendricks2016generating} defines a task similar to ours, to explain and justify a classification model. Their model is learned in a supervised manner, with explanations generated from an LSTM network which only implicitly depends on the internal feature maps.
It is essentially an image captioning task that generates captions with more class-discriminative information. 
Our method is unsupervised and does not rely on another black-box network to generate descriptions.


\textbf{Class activation map and network dissection  --}
Class Activation Map (CAM) identifies the most important region in an image by the linear combination of final conv-layer feature maps, whose weight is from the parameter in the fully connected layer that connects the feature map to the class label:
$M_c(x,y) = \sum_k w_k^c f_k(x,y)$,
where $M_c(x,y)$ measures the importance of spatial location $(x,y)$ for class $c$. $f_k(x,y)$ is the value at $(x,y)$ on the $k$th filter's feature map. $w_k^c$ is the weight that connected the $k$th feature map to the prediction class $c$. 

Network dissection~\cite{Bau_2017_CVPR} is perhaps the most similar work to ours. In~\cite{Bau_2017_CVPR}, a filter is associated to a concept by measuring the overlap between the thresholded filter feature map and the concept segmentation mask. Intersection over union $\text{IoU}_{k,c}$ is proposed to represent the accuracy of unit $k$ in detecting concept $c$. The main different between our work and network dissection is that we model the {filter} {attribute} relation as conditional multinomial probability distribution and propose a general Bayesian inference algorithm to link a filter to multiple attributes. Our algorithm relies only on image-level caption annotation instead of pixel-level segmentation annotation.

\section{Bayesian Inference Framework} \label{sec:algorithm}

For the filter-level interpretation, we seek to represent each network filter with its respective activation patterns in terms of visual attributes. Constructing a paired filter attribute dataset is unrealistic, because the filter (as a composite function) is not a well-defined concept with concrete examples. Instead, we propose leveraging off-the-shelf image caption annotations because they contain rich textual references to visual concepts. The intuition behind our filter-attribute association is simple: a filter can be represented by the images that strongly activate them and the visual attributes contained in such images should have a high probability of representing the filter. The joint consensus of all images in the dataset can increase the probability of the relevant visual attributes and suppress that of the irrelevant visual attributes. 


\subsection{Filter-Attribute Distribution}\label{sec:filter_apdf}

We denote $\mathcal{F} = \{f_i | i=1,...,m\}$ as the group of final conv-layer model filters.
We denote $\mathcal{X} = \{x_j | j=1,...,n\}$ as the set of input images. The filter $f$'s output for input $x$ is written as $f(x)$ (with a slight abuse of notations), which we call a feature map or filter activation. We consider models~\cite{He_2017_ICCV,Huang_2017_CVPR} with a global pooling layer $\phi$ followed by a single fully-connected layer. The global pooling layer output for $x$ is written as $\phi(f(x))$. The output of the fully-connected layer is the class prediction from $\mathcal{C} = \{c_k|k=1,...,o\}$. The weight matrix of the fully-connected layer is $W^{o \times m}$. A list of textual attributes from $\mathcal{T} = \{t_l|l=1,...,p\}$ is attached to each image. We loosely denote by $t \in x$ if $t$ is contained in $x$'s attribute list. $x^t$ represents  images that contain attribute $t$.

We're interested to know the representative visual attributes for a filter in the network's final-conv layer. For a given filter $f$, the probability that an attribute $t$ can represent its activation pattern is:
\begin{equation} \label{eq:1}
p(t|f) \propto p(f|t) p(t)
\end{equation}

$p(t)$ is the prior probability for visual attribute $t$. We consider the relative importance of attributes because attributes carry different amount of information. For example, ``small bird'' has less information than ``orange beak'' because the latter appears less in the text corpora and corresponds to a more important image feature. We employ the normalized TF/IDF feature as the attribute prior. The term frequency (TF) of a phrase is its number of occurrences in the same captioning file. The inverse document frequency (IDF) of a phrase is the logarithm of total file number divided by the number of files containing the phrase.

$p(f|t)$ measures the likelihood of attribute $t$ activating filter $f$. As attributes are not directly involved in the neural network, we introduce the input image as a hidden variable:
\begin{equation} \label{eq:2}
\begin{split}
p(f|t) &\propto p(f | \mathcal{X}, t) p(\mathcal{X}, t)\\
&= \prod_j  p(f | x_j, t) p(x_j, t) \\
\end{split}
\end{equation}
where $\mathcal{X}$ represents the set of input images. $p(x_j, t)$ measures the probability that $x_j$ contains $t$:
\begin{equation}\label{eq:3}
    p(x_j, t) = \left\{\begin{matrix}
1 & \text{if}\quad t \in x_j\\ 
0 & \text{otherwise.}
\end{matrix}\right.
\end{equation}

We use $x^{t}$ to represent images containing $t$ in their attribute list. According to Eqn~\ref{eq:2} and Eqn~\ref{eq:3}, images without attribute $t$ are zeroed out, so we have $p(f|t) \propto \sum_j p(f|x_j^{t}, t)$. $p(f|x_j^{t}, t)$ measures the likelihood that the image $x_j^{t}$ and the attribute $t$ are the reason for filter $f$'s activation. $f$ is conditionally independent of $t$ given $x_j^{t}$:
\begin{equation} \label{eq:4}
\begin{split}
p(f | x_j^{t}, t) & = p(f|x_j^{t}) \\
&\propto \phi(f(x_j^{t}))
\end{split}
\end{equation}
where $\phi(f(x_j^{t}))$ is the global pooling layer output for input $x_j^{t}$ and filter $f$, which measures how likely an image will activate a filter. To summarize, the posterior probability that attribute $t$ is the reason that filter $f$ activates is given by:

\begin{equation}\label{eq:5}
    p(t|f) \propto \text{TF/IDF}(t) \prod_j \phi(f(x_j^{t}))
\end{equation}

The approximation that $p(f | x_j^{t}, t) = p(f|x_j^{t})$ neglects the fact that when an image activates a filter, the feature map favors certain attributes over others. For example, if $f(x_j)$ highlights the head area of a bird, attributes related to ``head'', ``beak'' or ``eyes'' should be assigned with higher probabilities than attributes related to ``wings'' and ``feet''. Although this  approximation assigns equal probability to all visual attributes inside an image, it actually  works quite well in practice, as the joint consensus of all input images boosts true attributes and suppresses false ones. Note that the proposed method can easily adapt to datasets with  other forms of annotations like keypoints or part segmentation. Higher probability can be assigned to the visual attributes associated with a part when the feature activation map overlaps highly with its segmentation mask.


\begin{figure*}[t]
  
  \centering
  \includegraphics[width=0.95\linewidth]{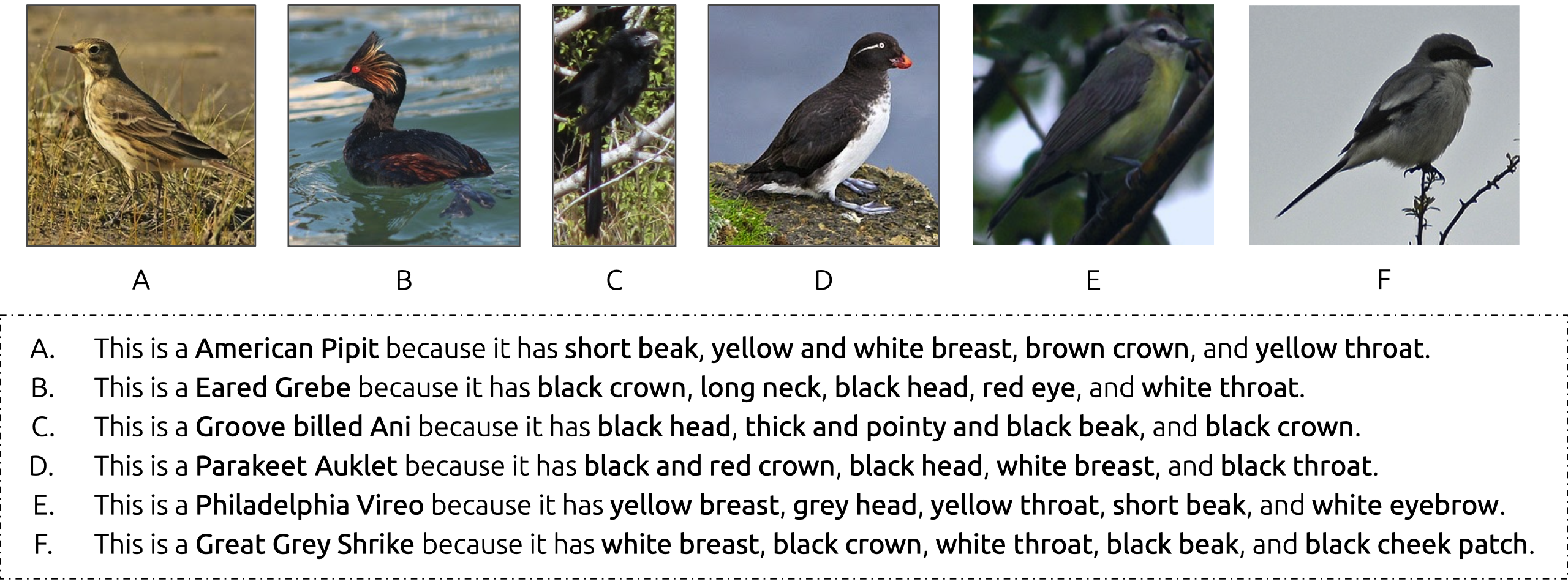}
  \caption[Textual summarization examples]{{Examples of textual summarization that contains the top visual attributes that are accountable for a network's decision-making. Note that the goal of textual summarization is not to accurately predict the image attributes, but to loyally reflect the reasons behind a neural network's classification decisions.}}
  \label{fig:image_text}
\end{figure*}

\subsection{Textual Summarization} \label{sec:summarize}
With the help of the filter-attribute distribution, we can find the top attributes that account for the network's classification decision. This task can be formulated as the probability that a visual attribute $t$ is the underlying reason given the fact that the network predicts input image $x$ as class $c$. We introduce final convolutional layer filters $\mathcal{F}$ as hidden variables and by marginalizing over $\mathcal{F}$ we get:

\begin{equation} \label{eq:6}
\begin{split}
p( t | x, c) &\propto p(t | \mathcal{F}, x, c) p(\mathcal{F} | x, c) \\
&=  p(t | \mathcal{F})  p(\mathcal{F} | x, c) \\
&= \prod_k p(t|f_i)  p(f_i | x, c)
\end{split}
\end{equation}
where $ p(t | x, c)$ is the probability that $t$ is the reason for the network predicting class $c$ for image $x$. $t$ is conditionally independent from $x$ and $c$ given $\mathcal{F}$, such that $p(t | \mathcal{F}, x, c) = p(t | \mathcal{F})$. $p(t | f_i)$ can be computed from filter $f_i$'s attribute distribution using Eqn~\ref{eq:5}. $p( f_i | x, c)$ measures the importance of filter $f_i$ in the decision-making process, and it's proportional to the product of the global pooling layer's output of $f_i$ denoted as $\phi(f_i(x))$ and the  weight between filter $f_i$ and class $c$, $w_{i,c}$: 
\begin{equation} \label{eq:7}
p( f_i | x, c) \propto \phi(f_i (x)) w_{i, c}
\end{equation}

We call $ p( \mathcal{T} | \mathcal{X}, \mathcal{C}) $ the class-attribute distribution.

A sentence is generated to describe the network's decision-making process using the class-attribute distribution. Although it's popular to employ a recurrent model for sentence generation, our task is to faithfully reflect the internal features learned by the network and introducing another network could result in additional uncertainty. We instead propose a simple template-based method using the top $n$ attributes, with the following form:

\vspace{0.05in}
\textit{"This is a \{class name\} because it has \{attribute 1\}, \{attribute 2\}, ..., and \{attribute n\}."}
\vspace{0.05in}

\begin{figure*}[ht!]
  
  \centering
  \includegraphics[width=1\linewidth]{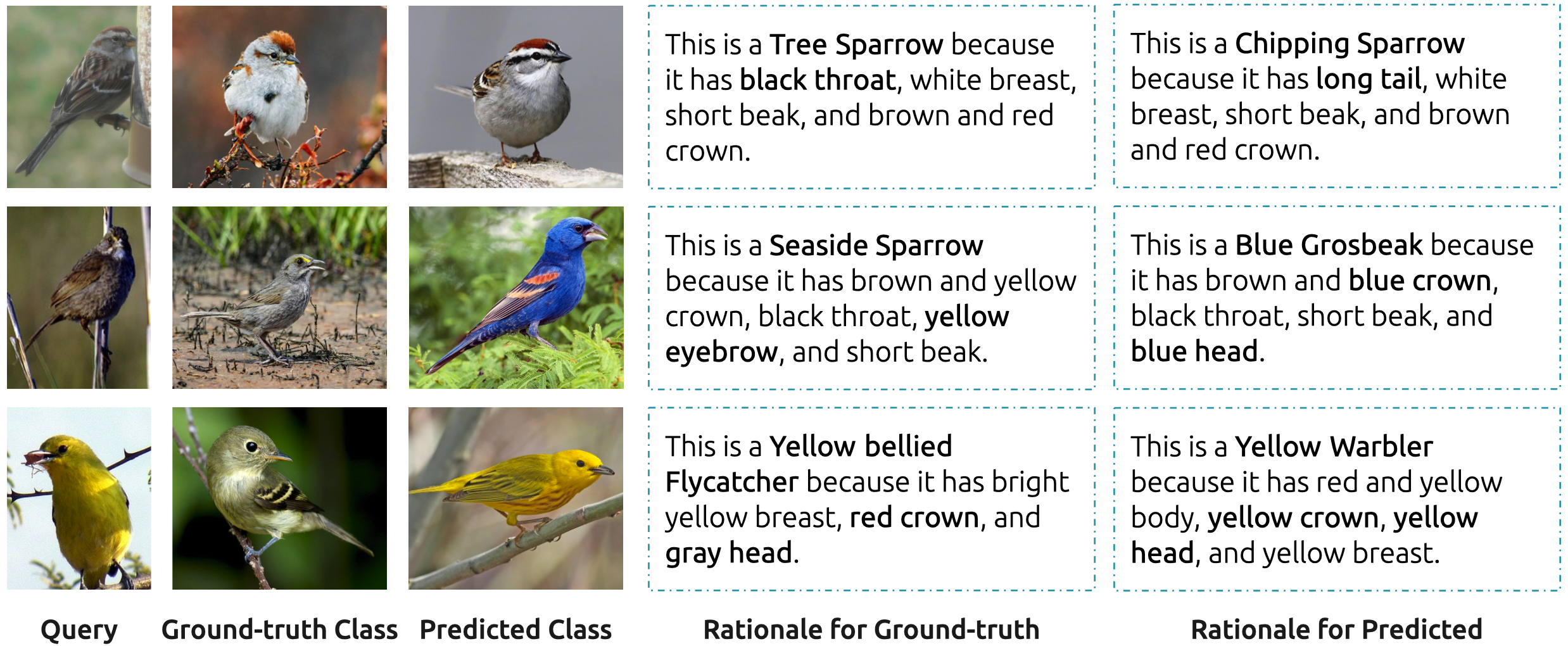}
  \caption[Network debugging examples]{Each row represents a network failure -- an incorrectly predicted class label.  From left to right, each column shows the query image, canonical images for both the ground-truth and the \emph{incorrectly} predicted classes, and the textual explanations for each of these classes.}
  \label{fig:debugging}
\end{figure*}

\section{Experiments}
\label{sec:experiments}

We evaluate the proposed algorithms on the fine-grained bird dataset of CUB-200-2011~\cite{WahCUB_200_2011} which contains 5997 training images and 5797 testing images. 
Two ways to obtain image-level attribute annotations for the CUB-200-2011 dataset are explored. The first is to leverage the image caption annotations provided by Zhang,~\etal~\cite{zhang2017stackgan}, which include five captions for every image that describes the visual features the  bird in the image has. Visual attributes are extracted from the captions as adjective-noun word phrases. The CUB-200-2011 dataset also provides visual attribute annotation. There are 312 total attributes to be labelled for each image. Examples include: ``Has bill length::longer than head'', ``Has back color::grey'' and ``Has back color::grey'', \etc
Although the visual attribute annotation can be more accurate, visual attributes from the captions are more diverse and fine-grained. All the visual attributes shown in this paper are generated from image captions.

To extract visual attributes from the image captions, we follow the process of word tokenization, part-of-speech tagging and noun-phrase chunking. A total of 9649 independent attributes are obtained. Examples of the generated filter-attribute distribution are shown in Figure \ref{fig:filter-att}. Examples of the generated textual explanations for image classification are shown in Figure~\ref{fig:image_text}.

\subsection{Network Debugging} 
\label{sec:debugging}
In figure~\ref{fig:debugging}, we show three major patterns of network failure through textual summarization. In the first example, a Tree Sparrow is incorrectly recognized as a Chipping Sparrow because the network mistakenly thinks ``long tail'' is a discriminative feature. According to wikipedia, American Tree Sparrows have a rufous stripe through the eye; on Chipping Sparrows it's black. Tree sparrows also have a spot in the middle of the breast and a bicolored bill that Chipping Sparrows don't have. Failing to identify the correct features for discrimination is the most common source of errors across the dataset. In fine-grained classification, the main challenge is to identify discriminative features for visually-similar classes, differences of which are often subtle and localized to small parts.

The second example shows a Seaside Sparrow that has mistakenly been recognized as a Blue Grosbeak.  From the textual explanations we ascertain that the low image quality mistakenly activates filters that correspond to blue head and blue crown. The underlying source of this error is complex -- the generalization ability of the network is limited such that small perturbations in the image can result in unwanted filter responses. Such failures imply the critical importance of improving network robustness to noisy inputs.

\begin{figure*}[ht]
  \centering
  \includegraphics[width=.98\linewidth]{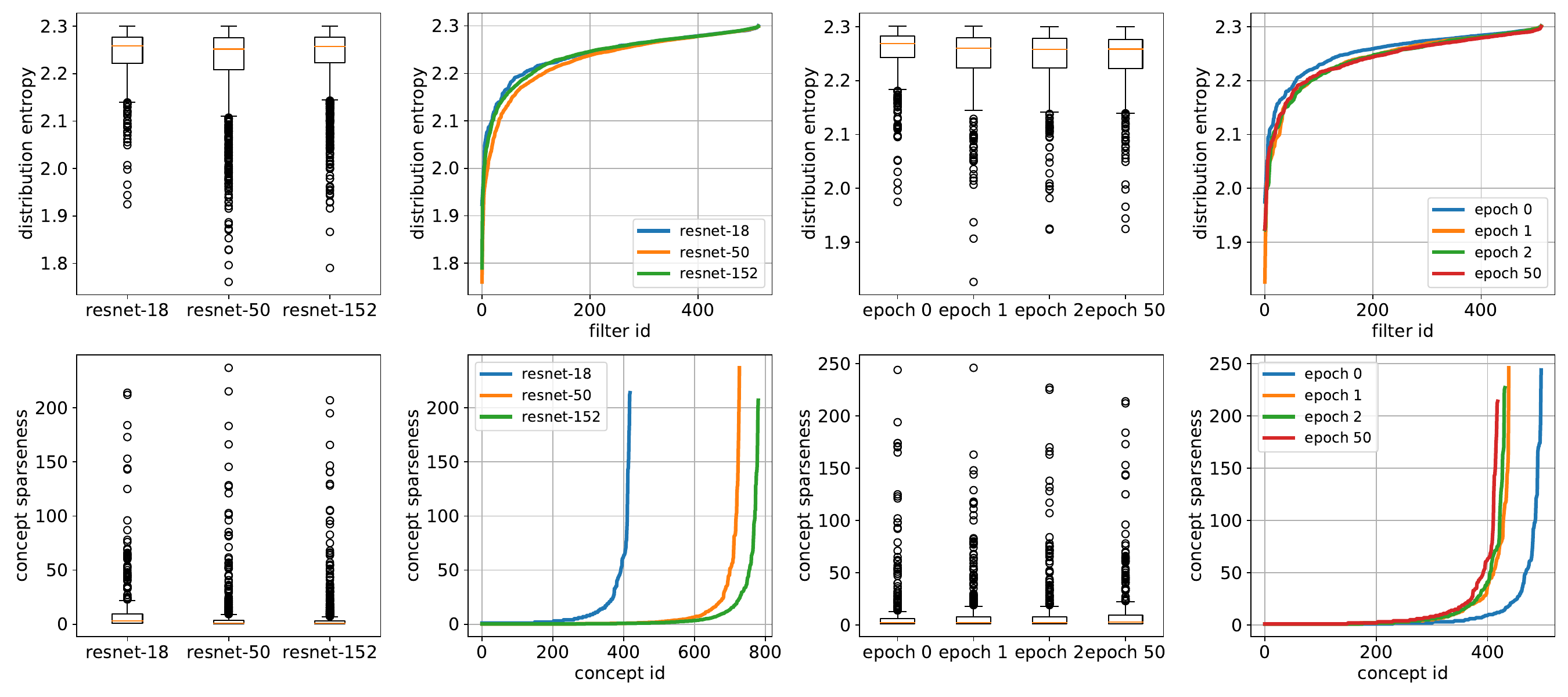}
\caption[Quantitative analysis]{Top Row: first two graphs shows the box plot and the sorted entropy for filter-attribute distribution of ResNet-18, ResNet-50, and ResNet-152 separately; last two graphs shows the filter-attribute distribution entropy of ResNet-18 at different training epochs of 0, 1, 2, 50. Botton Row: first two graphs shows the box plot and the sorted concept sparseness for Resnet-18, ResNet-50, and ResNet-152; last two graphs show the box plot and the sorted concept sparseness during training at epoch 0, 1, 2, 50 separately.}
\label{fig:network-epoch}
\end{figure*}

In the third case, the network predicts the image as a Yellow Warbler, however the ground-truth label is Yellow-bellied Flycatcher.  According to a bird expert, the network got this correct -- \emph{the ground-truth label is incorrect!}  The network correctly identifies the yellow crown and yellow head, both obvious features of the Yellow Warbler.  Errors like this are not surprising because, according to~\cite{van2015building}, the class labels on roughly 4\% of the CUB dataset are incorrect.

\subsection{Human Study}
\textbf{Visualization vs. semantic interpretation --}
We conduct human study using the Amazon Mechanical Turk platform to evaluate the proposed semantic interpretation. Our first study aims to know users' preference between visualization methods and semantic interpretation using textual summarization. "A picture is worth a thousand words". It's not surprising that normal users prefer the visualization method that have dominated the network attribution field. However, our study shows that 41.5\% of users prefer textual explanations and think they provide more helpful information than the visualization methods. This study confirms that semantic interpretation can serve as a helpful alternative or complement to network visualization.

\begin{table*}[ht]
\begin{center}
\begin{tabular}{|l | c | c | c | c | c | c |}
\hline
 & Mean & No.1 & No.2 & No.3 & No.4 & No.5  \\ 
\hline
Filter-attribute distribution & 80.1 & 93.9 &  92.9 & 89.8 & 74.3 & 53.2 \\
\hline
Class-attribution distribution& 75.8 & 89.2 & 88.1 & 83.3 & 66.4 & 51.9 \\
\hline
\end{tabular}
\end{center}
\vspace{-5pt}
\caption[Human study results]{The turkers rated accuracy for filter-attribute distribution and class-attribute distribution.}
\label{tab:branches}
\vspace{-5pt}
\end{table*}

\textbf{Filter-attribute distribution evaluation --}
In order to evaluate the filter-attribute distribution, we list the top five attributes along with the top-activated images for each filter. The users are instructed to select the attributes that are present in most, if not all the highlighted regions of these images. Generally 80.1\% of the attributes are regarded as accurate to describe the highlighted regions. Specifically, the accuracy for each of the five attributes are 93.9\%, 92.9\%, 89.8\%, 74.3\% and 53.2\%. This study shows that most of the visual attributes are rated as relevant to reflect the filter's activation pattern. To evaluate our textual summarization algorithm, we asked the turkers to rate the top five attributes for each image. The average accuracy was 75.8\%. The per-attribute accuracies were 89.2\%, 88.1\%, 83.3\%, 66.4\% and 51.9\%. Further study reveals another interesting phenomenon: for images that are correctly predicted by the network, 76.3\% of our top attributes were regarded as accurate; for incorrectly predicted images the accuracy was 73.8\%. This indicates the probability that a neural network learns better features for correct classification than incorrect classification.



\subsection{Network Understanding}
\label{sec:understanding}

If network \textit{interpretation} is about knowing what features a network had learned, network \textit{understanding} cares more about what makes a good feature. Defining good features and finding a way to learn them is crucial for the continuous success of deep models. Bengio, \etal~\cite{bengio2013representation} listed several characteristics a good feature representation should have, \textit{e.g.} disentangling factors of variation, smoothness, abstraction and invariance, and distributed representations. Many efforts~\cite{locatello2018challenging,higgins2018towards,szegedy2013intriguing,nguyen2015deep,ruderman2018pooling} have been devoted into understanding each of these properties. In this section, we focus on the distributed representation and its correlation with a model's performance.


\begin{figure*}[ht]
  \centering
  \begin{subfigure}[b]{0.55\textwidth}
  \includegraphics[width=1\linewidth]{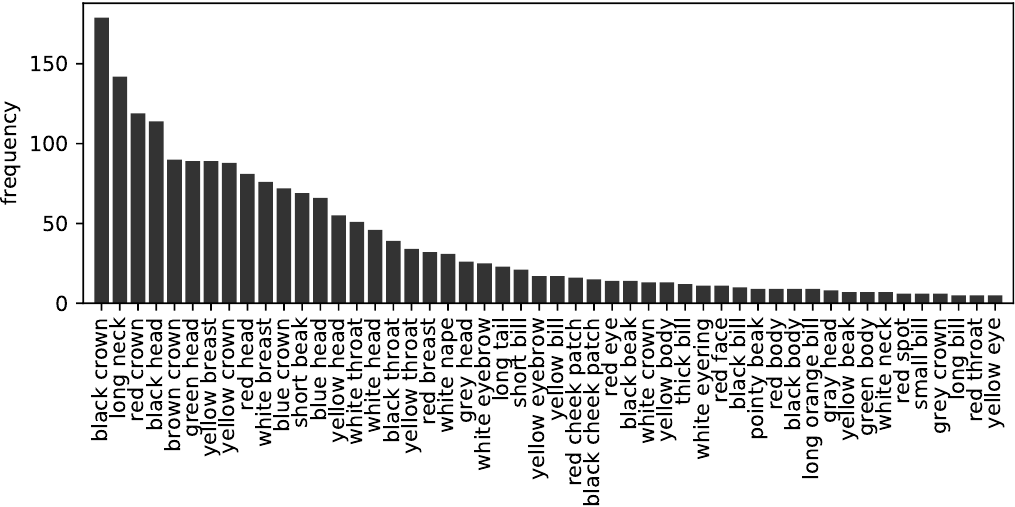}
  \caption{}
  \label{fig:att_dist}
  \end{subfigure}
  \begin{subfigure}[b]{0.42\textwidth}
  \includegraphics[width=1\linewidth]{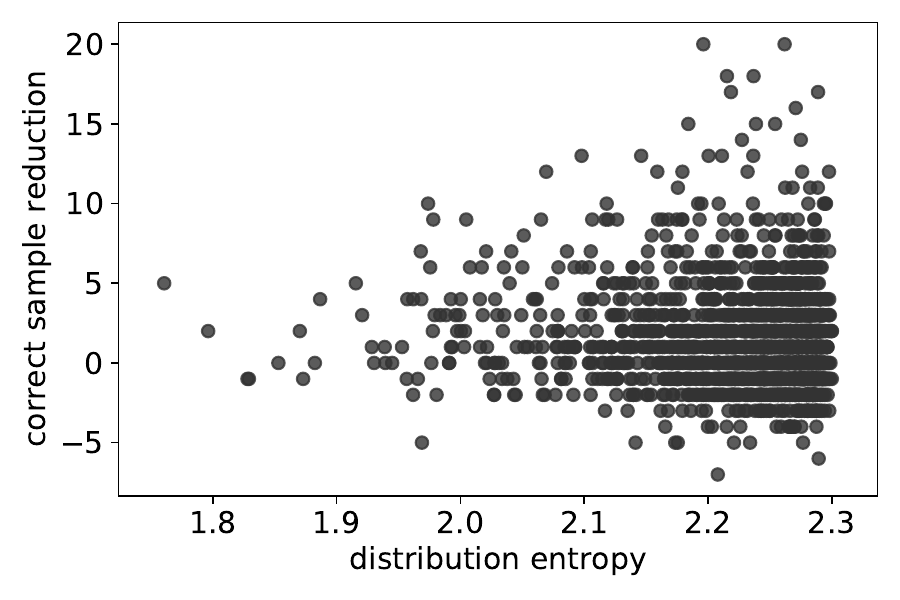}
  \caption{}
  \label{fig:scatter}
  \end{subfigure}
  \caption[Filter selectivity study]{(a) Visual attributes sorted by the number of filters that encodes them. (b) The scatter plot of filter selectivity (inversely proportional to its attribution distribution entropy) and filter importance (inversely proportional to correct sample reduction after removing it from the model).}
  \label{fig:att-scatter}
  \vspace{-5pt}
\end{figure*}

\textbf{Filter Selectivity --} 
A filter's selectivity refers to its representing a small number of concepts. A strongly-selective filter only activates on a narrow set of visual attributes. These filters are more interpretable than those whose activation patterns spread widely across many visual attributes. The entropy of the filter-attribute distribution serves as a good indicator of a filter's selectivity; low entropy means a sparse attribute distribution and strong selectivity. A few questions arise: do models with more interpretable filters perform better? Is the opposite true? Is there a significant correlation between a model's performance with its filters' selectivity? 

We compare the filter selectivity of three different models with increasing numbers of parameters: ResNet-18, ResNet-50, and ResNet-152. The classification accuracy on the CUB dataset for these models is 73.2\%, 81.6\% and 83.4\% respectively. The entropy of their filter-attribute distribution  are shown in Figure~\ref{fig:network-epoch}. Note that ResNet-18 has 512 filters and ResNet-50, ResNet-152 have 2048 filters. The first two graph on the top row of Figure~\ref{fig:network-epoch} show the box plot and sorted distribution entropy for the three models. ResNet-18 has the highest entropy and, by definition, lowest selectivity among all three models. ResNet-50 has more strongly-selective filters than ResNet-152 although the later is more accurate. To understand how filter selectivity evolves during training, we take four snapshots of a ResNet-18 network during training with epoch number 0, 1, 2 and 50. The box plot and sorted distribution entropy are shown in last two figures on the top row of Figure~\ref{fig:network-epoch}. Before training, the network has a lower number of selective filter, but the filter selectivity is not strictly increasing during training.

We next study the correlation between a filter's importance and its selectivity. A filter's importance can be measured by the performance drop after deleting it from the model: filters with higher correct sample reduction are of greater importance. We sequentially remove the final convolutional layer filters, one at a time, and record the decrease in correctly predicted samples. Note that the model is not retrained after filter removal. We compare the reduction of correct samples against the 
filter's selectivity in the scatter plot shown in Figure~\ref{fig:scatter}. Overall, deleting one filter typically has very little impact on the model's performance ($\pm5$), but note that the highest correct sample loss occurs when some of the most weakly-selective filters are deleted. Removing a strongly-selective filter is less likely to result in a performance drop compared to weakly-selective filters. This experiment shows that a filter's importance is surprisingly negatively related to its selectivity. We hypothesis that an important filter encodes some rare concepts and a less important filter encodes some concepts that are highly duplicate. We find that the concepts represented by the most strongly-selective filter are: ``yellow crown, black throat, black cheek patch'', which are encoded by many filters. Deleting such a filter is less likely to cause a significant dip in a model's performance. 

\textbf{Concept Sparseness --} Concept sparseness refers to the fact that a concept is represented by several filters. We represent a concept's sparseness by the number of filters whose top 10 activation pattern contains such a concept. Figure~\ref{fig:att_dist} shows the most popular concepts (visual attributes) in descending order. `black crown', the top concept, is spread across 179 filters, followed by the `long neck' concept spread across 149 filters. Note that `Black crown' is also the most frequent attribute in the caption annotation file. The bottom row of Figure~\ref{fig:network-epoch} shows how concepts are encoded in different models and how they changed during the process of training. ResNet-18 has less concepts encoded than ResNet-50, which is then followed by ResNet-152. During the training phrase for each model, the total concepts reduced but the number of filters that encode a concept increases. Generally speaking, better model encodes more concepts and the concepts become increasingly more distributed in the filters during training. 


\section{Conclusion}
\label{sec:conclusion}

In this paper, we focus on the task of semantic network interpretation at both filter and decision level. We represent the concepts a filter learns as a conditional multinomial probability distribution on visual attributes. A Bayesian inference algorithm is proposed to compute the attribute distribution for both filers and network decision. We study the correlation between a model's performance with its distributed representation. Two metrics (filter selectivity and concept sparseness) are examined. Generally, better models have higher filter selectivity and encode more concepts. During training, the filter selectivity increases and the concepts become increasingly more distributed in the filters. For decision-level semantic  interpretation,  textual summarization is generated to justify a network's classification results and can be used to uncover the common failure patterns on fine-grained recognition. Human studies are conducted to evaluate the accuracy of the proposed algorithms and validate the importance of semantic network interpretation.

{\small
\bibliographystyle{ieee_fullname}
\bibliography{interp}
}

\end{document}